\pdfoutput=1

\documentclass[letterpaper, 10 pt, conference]{ieeeconf}  

\IEEEoverridecommandlockouts                              

\usepackage{graphicx} 
\usepackage{cite}

\usepackage{mathptmx} 
\usepackage{times} 
\usepackage{amsmath} 
\usepackage{amssymb}  
\usepackage{xcolor} 
\usepackage{booktabs}
\usepackage{subcaption}
\usepackage{algorithm}
\usepackage{algpseudocode}
\usepackage{hyperref}

\title{\LARGE \bf
Fiber Optic Sensing Glove for High Performance Dexterous Manipulation Capture
}

\author{J.D. Peiffer$^{1,2,*}$, Taylor Niehues$^{1}$, Li Guan$^{1}$, Ziyi Kou$^{1}$, and Ergys Ristani$^{1}$
\thanks{*Work done at Meta}
\thanks{$^{1}$Meta,
        Redmond, Washington, 98052, USA
        {\tt\small ristani@meta.com}}%
\thanks{$^{2}$Department of Biomedical Engineering, Northwestern University,
        Evanston, Illinois, 60208, USA
        {\tt\small jpeiffer@sralab.org}}%
}

\begin{document}
\bstctlcite{IEEEexample:BSTcontrol}
\bibliographystyle{IEEEtran}

\maketitle
\thispagestyle{empty}
\pagestyle{empty}

\begin{abstract}

Capturing hand pose during dexterous manipulation remains difficult: vision-based methods degrade under occlusion and challenging lighting, while sensorized gloves, though occlusion-free, are prone to drift and magnetic interference and rarely match motion-capture accuracy. We introduce a fiber optic sensing glove for full hand pose tracking that targets these failure modes, using multi-core shape-sensing fibers that capture each fiber's full 3D shape rather than curvature alone. A novel pipeline registers each reconstructed fiber shape to a common hand reference frame, and a new inverse-kinematics solver reconstructs full hand pose at 60 Hz using curve constraints. Benchmarked on a 2-hour dataset of dexterous object manipulation tasks across 5 subjects, the glove achieves 7.2 mm mean fingertip position error against motion capture ground truth, reduced to 4.9 mm by a one-time factory calibration of the fiber routing hub that transfers across users and sessions. These capabilities enable high-fidelity data capture and bimanual virtual teleoperation — both essential to advancing the robotics field.

\end{abstract}

\section{INTRODUCTION}
Hands are central to how we interact with the physical and virtual world, and accurately tracking their motion is essential for analyzing dexterous behavior and teaching robots to perform the same tasks \cite{an_dexterous_2025,weinberg_survey_2024}. Yet the situations where tracking matters most during manipulation---fingers wrapped around an object or reaching into a confined space---are precisely where it is hardest. A decade of progress has produced large-scale hand pose datasets and a range of vision- and sensor-based systems, but each still trades off accuracy, robustness, portability, encumbrance, and cost.

Video-based hand tracking methods are popular and typically operate by estimating hand landmarks from a single camera \cite{qi_computer_2024,khirodkar_sapiens_2025}. Some are even able to run on resource-limited portable platforms such as smartphones and smart glasses \cite{zhang_mediapipe_2020}.  While recent years have seen an impressive increase in the accuracy of video-based systems, estimating hand pose from a single view during object manipulations is often ill-posed due to self-occlusion or object occlusion.

\begin{figure}
    \centering
    \includegraphics[width=\columnwidth]{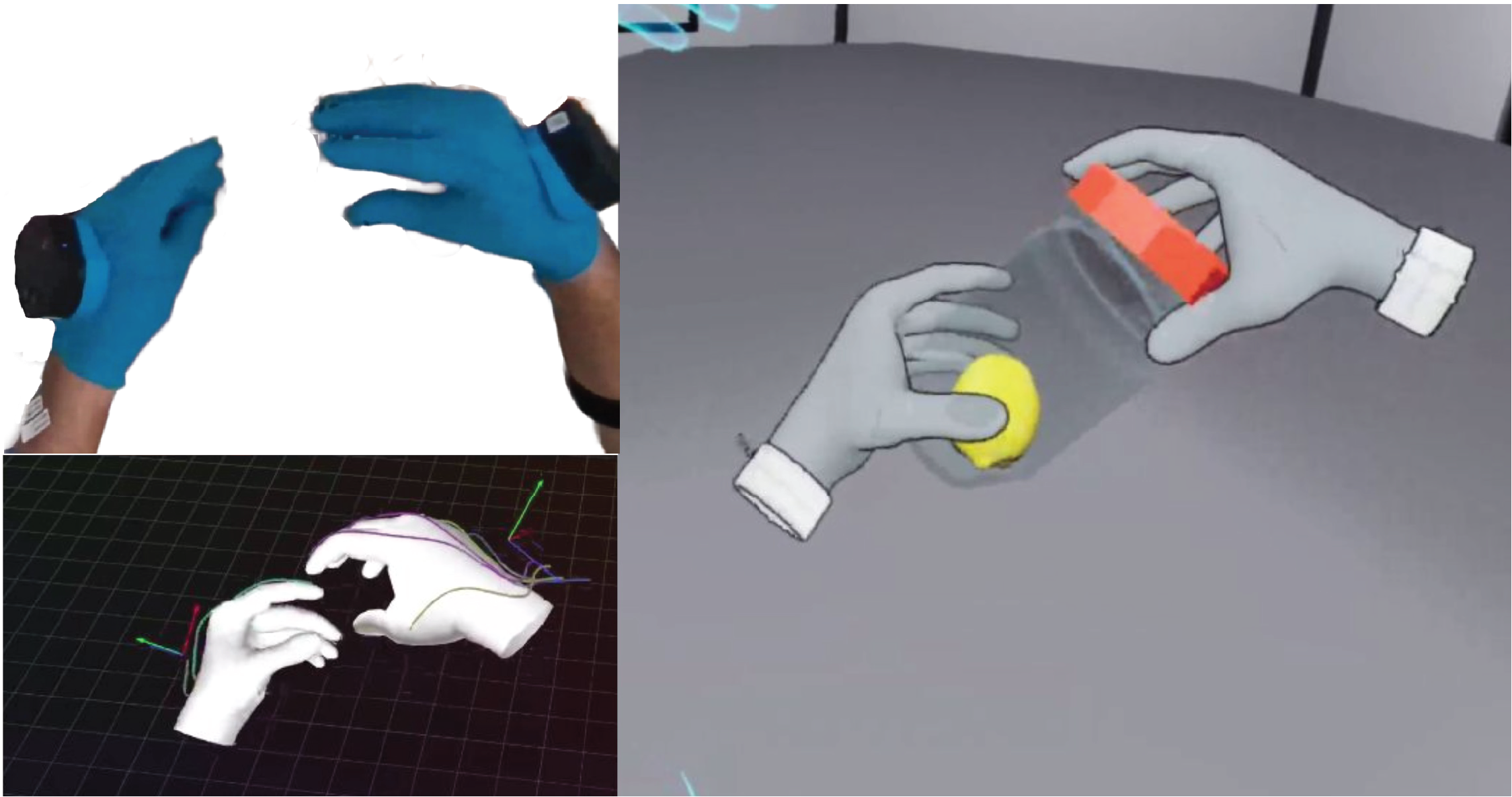}
    \caption{We demonstrate the utility of our gloves for occlusion-free, high accuracy dexterous manipulation data capture and for virtual teleoperation. \emph{Top left}: Our sensing gloves instrumented with multi-core optical fibers and a dense array of Fiber Bragg Gratings. \emph{Bottom left}: The system enables whole-hand pose estimation and is not vulnerable to drift, magnetic interference, or occlusion. \emph{Right}: The user successfully opens a jar in virtual reality with minimal effort.}
    \label{fig:splash}
    \vspace{-5mm}
\end{figure}

For higher accuracy and robustness to occlusion, multi-camera motion capture (mocap) systems are preferred. Marker-based systems achieve low marker position error and are commonly used as a benchmark in hand pose estimation \cite{han_online_2018}. Recent markerless multi-camera systems combine monocular pose estimates from multiple views and, by nature, allow unencumbered hand tracking \cite{firouzabadi_biomechanical_2024}. While accurate, these systems are limited to a calibrated capture volume and remain vulnerable to occlusion when hands reach into confined spaces like a pocket or box.

Instrumented gloves are frequently used when portability and continuous occlusion robustness are primary requirements \cite{lee_visual-inertial_2021,li_fsglove_2025,tchantchane_review_2023,caeiro-rodriguez_systematic_2021,pan_state---art_2023}. These systems span several sensing modalities, each with distinct strengths and limitations.
First, IMU-based gloves integrate multiple inertial measurement units to estimate segment orientations. While this approach is low-cost and scalable, it necessitates drift compensation \cite{sarker_real-time_2026}, with heading estimates being particularly susceptible to error accumulation.
Electromagnetic (EM) tracking gloves typically mount sensors on the fingernails, referenced to a base station located on the dorsum of the hand \cite{parizi_auraring_2020}. EM sensing can achieve high accuracy under controlled conditions but degrades in the presence of ferromagnetic materials, requiring careful management of the surrounding environment.
Strain and flex-sensor gloves infer joint motion from material deformation \cite{glauser_interactive_2019,park_stretchable_2024}. In practice, calibration remains challenging because sensor outputs are influenced by both bending and distributed strain.
Although sensorized gloves are inherently robust to visual occlusion, limitations in sensing accuracy and stability mean that no single modality has proven universally reliable across settings.

Fiber-optic bend sensors have been explored for hand tracking \cite{hu_fiber_2023}. They are attractive because they are immune to inertial drift, are largely unaffected by ferromagnetic disturbances, and they can be calibrated with relatively simple procedures. These gloves route thin optical fibers along the fingers and embed strain sensors---most commonly Fiber Bragg Gratings (FBGs)---whose local strain changes can be converted to curvature at discrete sensing locations.  Many prior designs attempt to keep each sensing region aligned with a specific joint so that measured curvature corresponds closely to the joint’s bend.

Most existing FBG gloves use single-core fibers and primarily estimate flexion at a small number of joints \cite{sun_wearable_2024,da_silva_fbg_2011}.
For example, Jha et al. \cite{jha_design_2021} uses FBG-derived curvature to estimate IMU-derived proximal interphalangeal (PIP) and metacarpophalangeal (MCP) flexion.
Rao et al. \cite{rao_study_2023} similarly estimate MCP, PIP, and distal interphalangeal (DIP) flexion.
Importantly, these systems do not recover full hand pose: a single-core FBG measures curvature magnitude but not bending direction, making it ill-posed to disentangle MCP flexion from adduction/abduction and other out-of-plane motions.

In contrast to existing work, we rely on multi-core FBGs, which measure both curvature magnitude and direction and thus enable reconstruction of a fiber's full 3D shape; because curvature is derived from differential strain between cores, common-mode effects such as temperature variation largely cancel, aiding stability over long captures.
Such shape-sensing fibers are well established in continuum and surgical robots \cite{floris2021fiber}, but they remain specialty components: our glove requires five fibers to be interrogated simultaneously at high sensor density, and interrogators offering this combination are not widely available as the fiber-optic industry has not previously targeted wearable applications.

The objective of this work is to develop a sensorized glove that tracks full hand pose with high accuracy while remaining robust to occlusion and drift during highly dexterous manipulation. To achieve this, we route multi-core fibers through stitched dorsal channels on a custom glove to minimize interference with natural motion, and use the resulting dynamic 3D fiber shapes as constraints for hand pose reconstruction. Our estimation pipeline first registers the reconstructed fiber shapes to the hand, then recovers full-hand pose with an inverse kinematics solver constrained by those shapes. We benchmark the glove on complex manipulation tasks, achieving less than 5 mm fingertip tracking error.

Finally, we demonstrate the system in a bimanual virtual teleoperation task, opening a virtual jar (Fig. \ref{fig:splash}).

In summary, our paper makes the following contributions.
\begin{itemize}
\item We design and demonstrate the viability of an optic fiber shape-sensing glove for whole-hand tracking.
\item We introduce a simple, effective hand-pose tracking pipeline based on shape registration, including an IK solver operating on spline constraints and a user self-calibration procedure.
\item We benchmark the system on representative manipulation tasks with competitive performance.
\item We demonstrate practical use, including hand-dexterity data capture and bimanual virtual teleoperation. 
\end{itemize}

\section{METHODS}
We first describe the glove hardware in Section~\ref{subsec:hardware}, followed by the hand model used for reconstruction in Section~\ref{subsec:hand_model}. We then present our method for estimating full hand pose from reconstructed fiber shapes in Section~\ref{subsec:reconstruction}, including the fiber-to-hub shape registration procedure and the inverse-kinematics solver. We describe our simple approach to self-calibration in Section \ref{subsec:self_cal}.

\subsection{Hardware}
\label{subsec:hardware}

\begin{figure}
    \centering
    \includegraphics[width=\columnwidth]{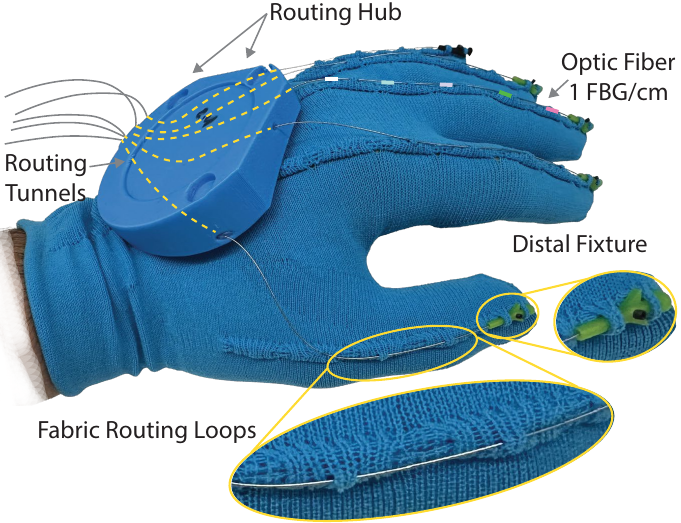}
    \caption{Our sensing glove routes multicore optical fibers---which reconstruct 3D shape---through fabric loops along each finger. The fibers are fixed at the fingertip but otherwise allowed to move freely along the fabric loops and through the routing hub. The routing hub uses known tunnel geometry to register 3D fiber shape reconstructions to a common reference frame.}
    \label{fig:hardware}
    \vspace{-5mm}
\end{figure}

\begin{figure*}
    \centering
    \includegraphics[width=\textwidth]{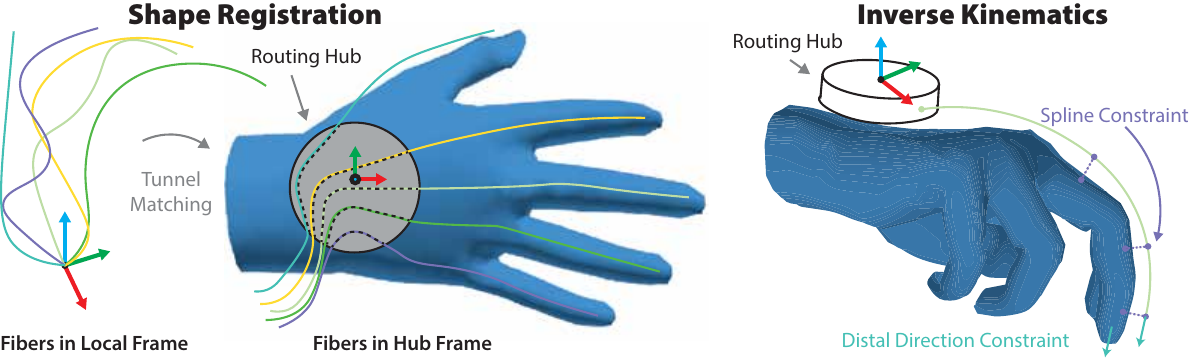}
    \caption{Optic fibers are first reconstructed in their own reference frame. The shape registration process matches routing hub tunnels to the section of the fiber with same curvature. This is done for every fiber at every time sample to put the fibers in the routing hub frame. Next, inverse kinematic solutions uses spline position and direction costs to pose the hand model.}
    \label{fig:pose_reconstruction}
    \vspace{-5mm}
\end{figure*}

\subsubsection{Glove}
The glove consists of the glove fabric, optical fibers, fixtures, and a fiber routing hub (Fig. \ref{fig:hardware}). The fabric is custom-knit for comfort and sized for large hands. Each finger is instrumented with one optical fiber which is rigidly attached on a 3D printed fixture near the fingernail. The fiber is routed along the dorsal surface of each finger through dorsal fabric loops and allowed to move along the length of the finger. Each fiber is embedded in a nitinol tube with 0.4 mm outer diameter for protection. The dorsal fabric loops help prevent overly tight bend radii and keep the fiber medially on the finger. At the dorsum of the hand, fibers enter a 3D printed dorsal routing hub glued to the glove. Within the hub, each fiber enters near the MCP joint at the base of its finger and is guided across the dorsum to a shared ulnar (pinky-side) outlet. The hub’s internal tunnel geometry is used for shape registration (Section \ref{subsec:shape_registration}). Optionally, a 6 DoF tracking system is mounted on the dorsal routing hub to provide world tracking for the hand.

\subsubsection{Fiber Optic Sensors}
Each optical fiber has 125 um diameter and consists of three sensing cores, with each core containing 26 FBG sensors spaced at 1 cm intervals, beginning 1 cm from the distal fiber tip. FBG wavelengths are spaced 3 nm apart, spanning between 1511-1587 nm. The sensors are entirely passive, the light emission and detection are all done remotely on a FBG interrogator device. We used the ShapeScan 905 interrogator from FBGS\footnote{\url{https://www.fbgs.com/}} (Jena, Germany) which is tailored for Wavelength Division Multiplexing (WDM) sensing across 5 multi-core fibers. The interrogator is placed on a desk or rolling cart for mobility. It connects to the sensing fibers on the glove via 5 meter long patch cables at a wrist interconnect.

\subsubsection{Fiber Shape Reconstruction}
Each individual FBG sensor measures wavelength shift which can be mapped to local strain using the fiber core geometry. By combining strain measurements from 3 cores at a sensing location, local curvature (mm$^{-1}$) and bending direction (rad) can be measured precisely. There are several choices for integrating these local measurements on a curve to obtain shape, including the Frenet-Serret formulation\cite{moore_shape_2012}, the Bishop frame method\cite{bishop1975there}, or deep learning\cite{ManaviRoodsari2024}. In this work, we used the proprietary algorithm included in the ShapeScan 905 device with 1 mm spatial resolution. This produced a time-varying shape representation $\mathbf x \in \mathbb{R}^{251 \times 3}$ for each fiber. We sampled curvature, angle, and shape estimates for each fiber at 60 Hz.

\subsection{Hand Model}
\label{subsec:hand_model}
We model the hand as a kinematic tree parameterized by wrist rotation, wrist translation, and 22 joint angles (5 $\times$ 2 for MCP, 5 $\times$ 1 for DIP and 5 $\times$ 1 for PIP and 2 for wrist) using the Momentum Library \cite{noauthor_facebookresearchmomentum_2026,ferguson_mhr_2025}. This model contains 21 anatomical landmarks: the wrist and palm center, and for each digit the proximal, intermediate, and distal phalangeal landmarks plus the fingertip landmark. This model is scaled before tracking using a pre-measured hand size. Model skinning is performed for visualization.

\subsection{Hand Pose Reconstruction}
\label{subsec:reconstruction}

\subsubsection{Fiber to Routing Hub Registration}
\label{subsec:shape_registration}
Fiber shape reconstructions $\mathbf x$ are initially expressed in a local frame relative to the first FBG sensor (Fig. \ref{fig:pose_reconstruction}). To express them in the common routing hub frame, we register each reconstructed fiber curve to a pre-measured hub tunnel centerline for that digit (Algorithm \ref{algo:shape_registration}).

\begin{algorithm}[H]
\caption{Fiber to Routing Hub Shape Registration}
\label{algo:shape_registration}
\begin{algorithmic}[1]
\Require Routing Hub Tunnel Points $\mathbf{T}$, Fiber Points $\mathbf{F}$
\Ensure Best rigid transform $(\mathbf{R}^\star,\mathbf{p}^\star)$
\State $\kappa_T \gets \mathrm{Curvature}(\mathbf{T})$
\State $\kappa_F \gets \mathrm{Curvature}(\mathbf{F})$
\For{$j=1$ to $|\mathbf{F}|$}
  \State $c_j \gets \mathrm{RMSE}(\kappa_F[j:j+|\mathbf{T}|], \kappa_T)$
\EndFor
\State $\mathcal{M} \gets \mathrm{RelativeMinima}(\{c_j\})$
\State $j^\star \gets \arg\min_{j \in \mathcal{M}} \ \mathrm{RMSE}\big(\mathrm{Align}(\mathbf{T}, \mathbf{F}[j:j+|\mathbf{T}|])\big)$
\State $(\mathbf{R}^\star,\mathbf{p}^\star) \gets \mathrm{Align}(\mathbf{T}, \mathbf{F}[j^\star:j^\star+|\mathbf{T}|])$
\State \Return $(\mathbf{R}^\star,\mathbf{p}^\star)$
\end{algorithmic}
\end{algorithm}
\vspace{-3mm}

To find the alignment, we search over a 1D sliding-window index $j$ that selects a contiguous segment of the measured curve of the same length as the hub tunnel. For each candidate index in a digit-specific search range, we compute a curvature matching cost between $\kappa_{F}$ from the measured fiber $\mathbf x$ and $\kappa_{T}$ from the routing hub tunnels
where $\kappa = 1/R$ is the discrete curvature estimated from circumcircles of consecutive point triplets (endpoints omitted). To improve temporal stability, the search range is narrowed to a window around the previous frame's best index when available.

We first evaluate the curvature-matching cost $c(j)$ for each integer offset $j$ within the digit-specific search range. We then extract a set of candidate offsets $\mathcal{M}$ given by the relative minima of $c(j)$. For each $j\in\mathcal{M}$, we compute the optimal rigid transform $(\mathbf{R}_j,\mathbf{t}_j)$ aligning the measured segment to the template using the closed-form SVD-based Procrustes solution, and score it by the post-alignment sum of squared point-to-point residuals. We then return the transform with the lowest residual. This registration is performed independently for each digit at every time step. If a large timestamp gap ($>0.5$\,s) occurs, the tracker resets its temporal state (the previous best offset).

We upsample both the hub tunnel curve and the measured fiber shape using a cubic B-spline fit and re-sampling with an integer multiplier (default 3) for the registration process, however return the non-upsampled fiber shapes. This was done to prevent jitter induced by the registration jumping between point indices $j$.

\subsubsection{Inverse Kinematic Solution}
\label{subsec:ik}
For each frame, we estimate a full hand pose by fitting a scaled Momentum Hand Model \cite{noauthor_facebookresearchmomentum_2026,ferguson_mhr_2025} to the reconstructed 3D optical-fiber shapes. The model is instantiated with a subject-specific hand scale (measured prior to reconstruction) and augmented with per-digit landmark locators placed on the distal (fingernail), intermediate, and (for non-thumb digits) proximal phalanges. Let $\theta$ denote the vector of model joint parameters. We solve for $\theta$ by minimizing a weighted sum of geometric residuals constructed from the reconstructed 3D fiber points (Fig. \ref{fig:pose_reconstruction}).

\paragraph{Fingertip position and distal direction}
We constrain the distal locator position to match the spline endpoint (position residual), and we constrain the distal anatomical axis of the distal phalanx to align with $\hat{\mathbf{d}}$ (fixed-axis residual). These terms encourage both correct fingertip placement and plausible distal phalanx orientation.

A distal direction target is computed from the spline as the vector between the endpoint and a point near the end of the spline (the last 5 point samples), i.e.,
$\hat{\mathbf{d}}=\frac{\mathbf{x}_{N}-\mathbf{x}_{N-5}}{\|\mathbf{x}_{N}-\mathbf{x}_{N-5}\|}$.

\paragraph{Spline Cost}
To encourage the finger to follow the reconstructed fiber, we add point-to-spline constraints at the intermediate locator for every digit and at the proximal locator for the four fingers. For a locator point $\mathbf{y}(\theta)$ and discrete spline points $\{\mathbf{s}_j\}$, the residual is
\begin{equation}
\mathbf{r}_{\mathrm{spline}} = \mathbf{y}(\theta) - \mathbf{s}_{j^\ast},
\end{equation}
where
\begin{equation}
j^\ast = \arg\min_j \left\|\mathbf{y}(\theta) - \mathbf{s}_j\right\|^2
\end{equation}
is the nearest point on the spline. Nearest-neighbor association is recomputed at each evaluation.

\paragraph{Dorsum Cost}
Before tracking, we calibrate the position of the routing hub with respect to the hand mesh (described in Section \ref{subsec:self_cal}). During tracking we enforce (i) an orientation constraint on the dorsum frame, (ii) a position constraint, and (iii) a planar constraint that fixes the dorsum point to a plane with a hard-coded dorsal normal direction.

\paragraph{Temporal regularization}
To reduce frame-to-frame jitter, we optionally add an $\ell_2$ penalty on the model parameters centered at the previous frame's solution (used when a recent pose exists and the timestamp gap is small).

\paragraph{Objective and optimization}
All constraints are implemented as Momentum skeleton error functions (position, orientation, plane, fixed-axis angle, and point-spline position), and we additionally include strong joint-limit penalties to discourage anatomically invalid solutions. The per-frame objective is a weighted sum of these terms:
\begin{equation}
\begin{aligned}
L(\theta) =\;&
\lambda_{distal} L_{distal} +
\lambda_{spline} L_{spline} +
\lambda_{dorsum} L_{dorsum} \\
&+
\lambda_{prev} L_{prev} +
\lambda_{lim} L_{lim}.
\end{aligned}
\end{equation}
We use the default weights from our implementation:
$\lambda_{distal}=1$ (distal-axis),
$\lambda_{spline}=0.5$ (point-spline),
$\lambda_{dorsum}$ uses $\lambda_{dorsum}^{ori}=10$ and $\lambda_{dorsum}^{pos}=1$ (with the position term scaled by $0.01$ in the position constraint),
$\lambda_{prev}=4$ when enabled,
and $\lambda_{lim}=100$.

For each frame, we initialize the solver either from the previous frame’s pose (when available) or from a neutral hand pose. We then minimize $L(\theta)$ using a Gauss--Newton solver (Momentum \texttt{GaussNewtonSolver}) \cite{noauthor_facebookresearchmomentum_2026} for a fixed number of iterations, and update the hand model state with the optimized parameters. In addition, if an external root transform is provided (e.g., from a tracked dorsum reference), we apply it by aligning the model’s dorsum locator to the measured transform after IK, yielding the final posed hand in world coordinates.

\subsection{Routing Hub Self-Calibration}
\label{subsec:self_cal}

We calibrate the position and orientation of the modeled routing hub using a short time sequence of recorded samples. For each sample, we first run a standard per-frame IK solve with dorsum constraints disabled to obtain a reasonable initialization for the hand parameters, and we store these parameters in the sequence state. We then enable dorsum constraints and add the usual IK error terms for that frame (including dorsum position/orientation/plane terms) to the sequence objective. The resulting problem is solved with a Gauss--Newton sequence solver.

\section{Experiments}
\subsection{Experimental Setup}
We demonstrate our glove's utility in dexterous manipulation data capture by collecting a dataset of five participants performing varied hand poses and object interactions. Each participant completed two sessions, with a full glove doff-and-don between them to capture re-donning variability.

Each session comprised two phases: range of motion followed by manipulation. During the range of motion phase, participants executed poses designed to span common finger–finger contact configurations, including pinch, palm flat on the table, and a thumb-to-finger side-rub. During manipulation, participants completed a set of functional tasks: buzz wire, box and blocks, cup stacking, in hand object rotation, 9-Hole peg test, and object squeezing. In total, this amounted to 38 minutes in the range of motion phase and 79 minutes in the manipulation phase, for $\sim$2 hours of movement data.

In each session, participants wore the glove instrumented with optic fibers, along with 26 mocap markers for ground-truth tracking. Of these markers, 19 were used for hand pose estimation---three markers per finger, four on thumb and three on dorsum---and 7 were placed on the routing hub at the tunnel entrances/exits. The markers on each finger were placed on top of fiber loops. We estimated ground-truth hand pose from the mocap markers using the approach described by Han et al. \cite{han_online_2018}, with 20 calibrated OptiTrack cameras (NaturalPoint, Inc) surrounding the participant.

\subsection{Evaluation Methodology}
\begin{figure}
  \includegraphics[width=\columnwidth]{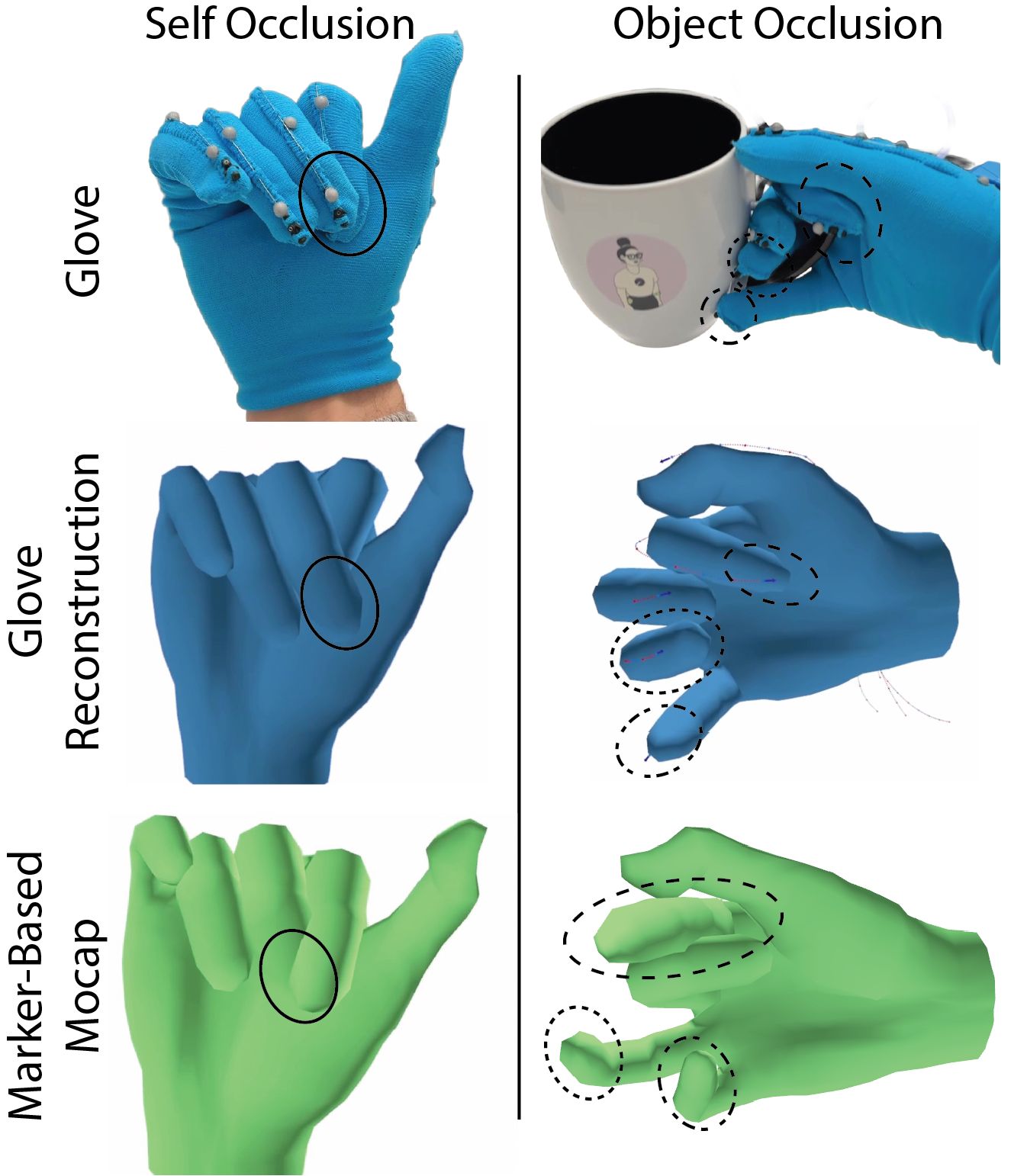}
  \caption{Our system allows for tracking of hand pose in situations where self-occlusion or object occlusion prevent marker-based systems from accurately tracking the whole hand. \emph{Left}: Self occlusion causes mocap to cross fingers incorrectly \emph{Right}: Marker occlusion on index, ring, and pinky fingers results in incorrect fingertip position. }
  \label{fig:occlusion}
  \vspace{-5mm}
\end{figure}
We report two types of errors: (1) fiber reconstruction/registration accuracy, measured against mocap markers placed on routing loops, and (2) hand landmark Euclidean error (mm) between our fiber-based IK reconstruction and mocap-based hand reconstruction. As mocap markers can be occluded (Fig. \ref{fig:occlusion}), we evaluate only frames in which all mocap markers on a finger are detected. We align glove and mocap reference frame using a per-frame rigid transform estimated from markers placed at the routing hub tunnel entrances/exits.

\paragraph{Fiber-to-marker Error}
To quantify 3D errors after fiber reconstruction and tunnel registration, we calculate the Euclidean distance from each finger's mocap markers to the closest point on its 3D fiber. We averaged these distances over markers on each finger and then over all fingers; we refer to this metric as \emph{fiber-to-marker error}. As it isolates errors prior to model fitting, it can be interpreted as the combined fiber-reconstruction and registration error.

\begin{figure*}
    \centering
    \includegraphics[width=\textwidth]{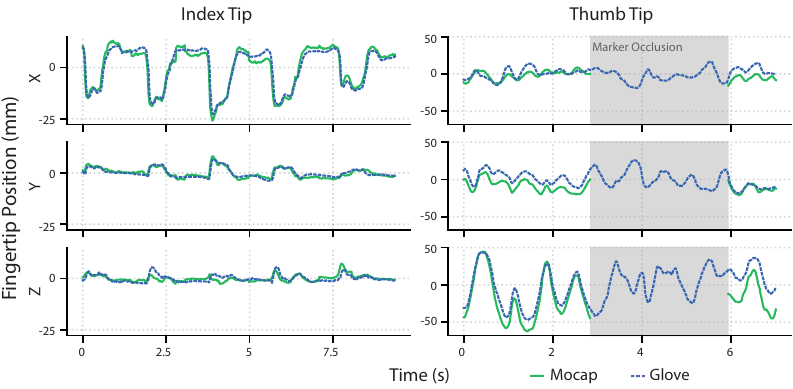}
    \caption{Our method tracks fingertip landmarks closely across a variety of manipulation tasks (Left: 9 Hole Peg Test, Right: in hand object rotation). Notably, our glove tracks during periods of mocap marker occlusion.}
    \label{fig:traces}
    \vspace{-5mm}
\end{figure*}

\paragraph{Hand Keypoint Error}
To evaluate whole-hand model reconstruction quality, we computed the Mean Keypoint Position Error (MKPE) between the glove-estimated hand pose and the mocap-derived hand pose, both represented using the Momentum hand model. MKPE is the mean Euclidean distance over 21 model landmarks, and Fingertip MKPE (F.MKPE) is computed over fingertip landmarks only. Qualitative traces of fingertip errors are shown in Fig. \ref{fig:traces}.

\subsection{Routing Hub Tunnel Alignment}
\label{subsec:optitrack_cal}
\begin{figure}
    \centering
    \includegraphics[width=\columnwidth]{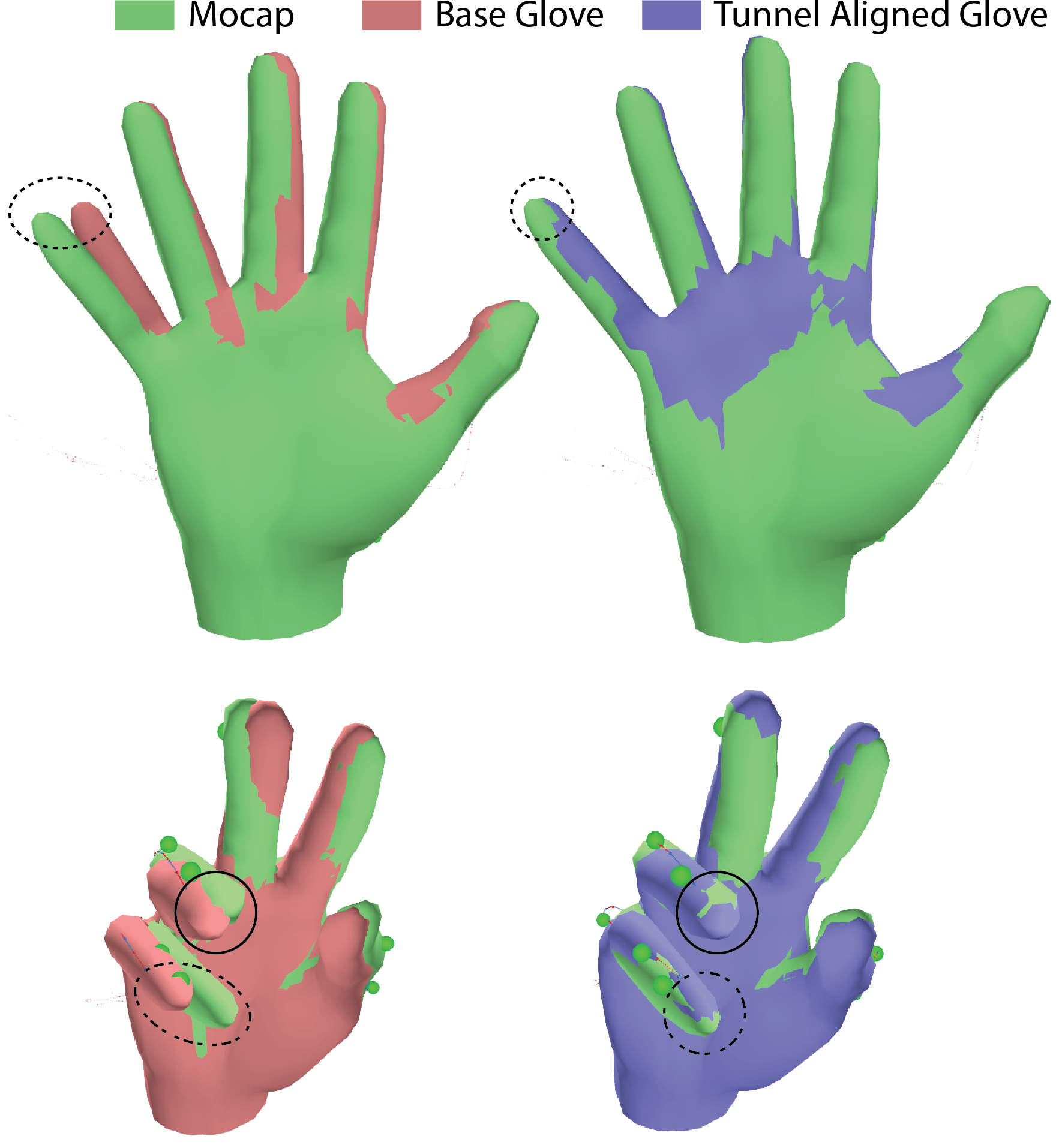}
    \caption{Un-aligned glove pose estimates (red) are corrected with a one-time routing tunnel alignment to repose fibers and hand mesh (blue) to better align with ground truth (green). First row: a bias in the rotation of the fiber is corrected to align the pinky. Second row: alignment brings the ring finger closer to the thumb in a pinch.}
    \label{fig:calibration}
    \vspace{-5mm}
\end{figure}
Errors in fiber shape reconstruction can propagate through our pipeline, leading to incorrect tunnel registration (Section~\ref{subsec:shape_registration}) and, in turn, degraded inverse kinematics estimates (Section~\ref{subsec:ik}). 

If these discrepancies are dominated by a consistent spatial bias (e.g., a systematic offset or rotation), they should be correctable with a single rigid transform. We therefore estimated the optimal rigid transform between the registered fiber curves and the mocap marker positions during the range of motion portion of our protocol across all frames and poses. We then applied this transform to the routing hub tunnel templates and re-ran the shape registration step using the corrected tunnel geometry, yielding bias-corrected registered fiber curves for subsequent IK. Qualitative results of this tunnel alignment are shown in Figure \ref{fig:calibration}.

\subsection{Quantitative Results}
\subsubsection{Fiber-to-marker Errors}

Average fiber-to-marker errors are shown in Table \ref{tab:benchmark}. Overall errors were low: 3.3 and 3.4 mm during range-of-motion and manipulation, respectively, indicating accurate shape reconstruction and registration. Tunnel alignment reduced errors by 1.2–1.5 mm, with a larger decrease during range of motion, where alignment was estimated. This correction generalized to manipulation, reducing fiber-to-marker error from 3.4 to 2.1 mm.

\subsubsection{Hand Keypoint Error}
In general, hand keypoint errors between mocap and glove hand model reconstructions were low, averaging 6.5 to 7.2 mm F.MKPE before tunnel alignment and 4.5-4.9 mm F.MKPE in both phases after alignment (Table \ref{tab:benchmark}). As expected, the errors at the fingertips (F.MKPE) were higher than errors averaged across the whole hand (MKPE). Tunnel alignment decreased error in every subject and session (Fig. \ref{fig:histogram}). Per-fingertip analysis found the thumb to have the highest error at 7.2 mm after alignment (Table \ref{tab:per_finger}). The pinky fingertip initially had the highest error before alignment (9.4 mm) but was brought down to 6.2 mm following alignment (Fig. \ref{fig:calibration}). The higher error is due to the pinky tunnel having the shortest length, affecting fiber-to-tunnel registration.

\subsubsection{Runtime}

The ShapeScan 905 device acquires and reconstructs a sample of 5 fibers shapes at 60\,Hz with a cumulative latency of 32.5\,ms. We benchmarked the downstream pipeline on a desktop with an AMD Ryzen Threadripper PRO 5975WX. Shape registration---implemented in Python---ran at 5.3\,ms/sample with 3$\times$ upsampling and 3.2\,ms/sample without, while inverse kinematics---implemented in C++ via the Momentum library~\cite{noauthor_facebookresearchmomentum_2026}---took 0.12\,ms/sample, well within the 16.7\,ms budget for 60\,Hz.

\subsection{Occlusion}
During the range of motion phase, 25\% of the mocap hand tracking samples had at least one marker occluded. During the manipulation phase 34\% of samples were occluded. 

We present examples of self and object occlusion in Figures \ref{fig:occlusion} and \ref{fig:traces}, where our glove produces visually consistent hand poses while marker-based mocap produces implausible poses. Importantly, these poses are not very extreme; simply grasping a coffee cup handle causes mocap to fail.

\subsection{Application Demonstration}
\label{subsec:bimanual}
To enable world-space tracking, we attach a 6-DoF tracker to the routing hub and use the resulting root transforms to place both hands in a shared world coordinate frame. Combined in real-time with our glove's finger pose estimates, this setup supports bimanual object interaction in virtual reality (Fig. \ref{fig:splash}). A full video of this teleoperation is shown in the supplementary video.

In this demo, the participant was able to open and close a virtual jar, demonstrating our system can be used for real-time virtual teleoperation.

\begin{table*}
\caption{Pose Benchmarking Errors}
\label{tab:benchmark}
\centering
\begin{tabular}{lcccccc}
\toprule
Metric & \multicolumn{2}{c}{Fiber-to-Marker Residual (mm)} & \multicolumn{2}{c}{MKPE (mm)} & \multicolumn{2}{c}{F.MKPE (mm)} \\
Phase & Range of Motion & Manipulation & Range of Motion & Manipulation & Range of Motion & Manipulation \\
\midrule
Base & 3.3 & 3.4 & 5.3 & 5.9 & 6.5 & 7.2 \\
Tunnel Aligned & 1.7 & 2.1 & 4.4 & 4.4 & 4.5 & 4.9 \\
\bottomrule
\end{tabular}
\vspace{-5mm}
\end{table*}

\begin{figure}
    \centering
    \includegraphics[width=\columnwidth]{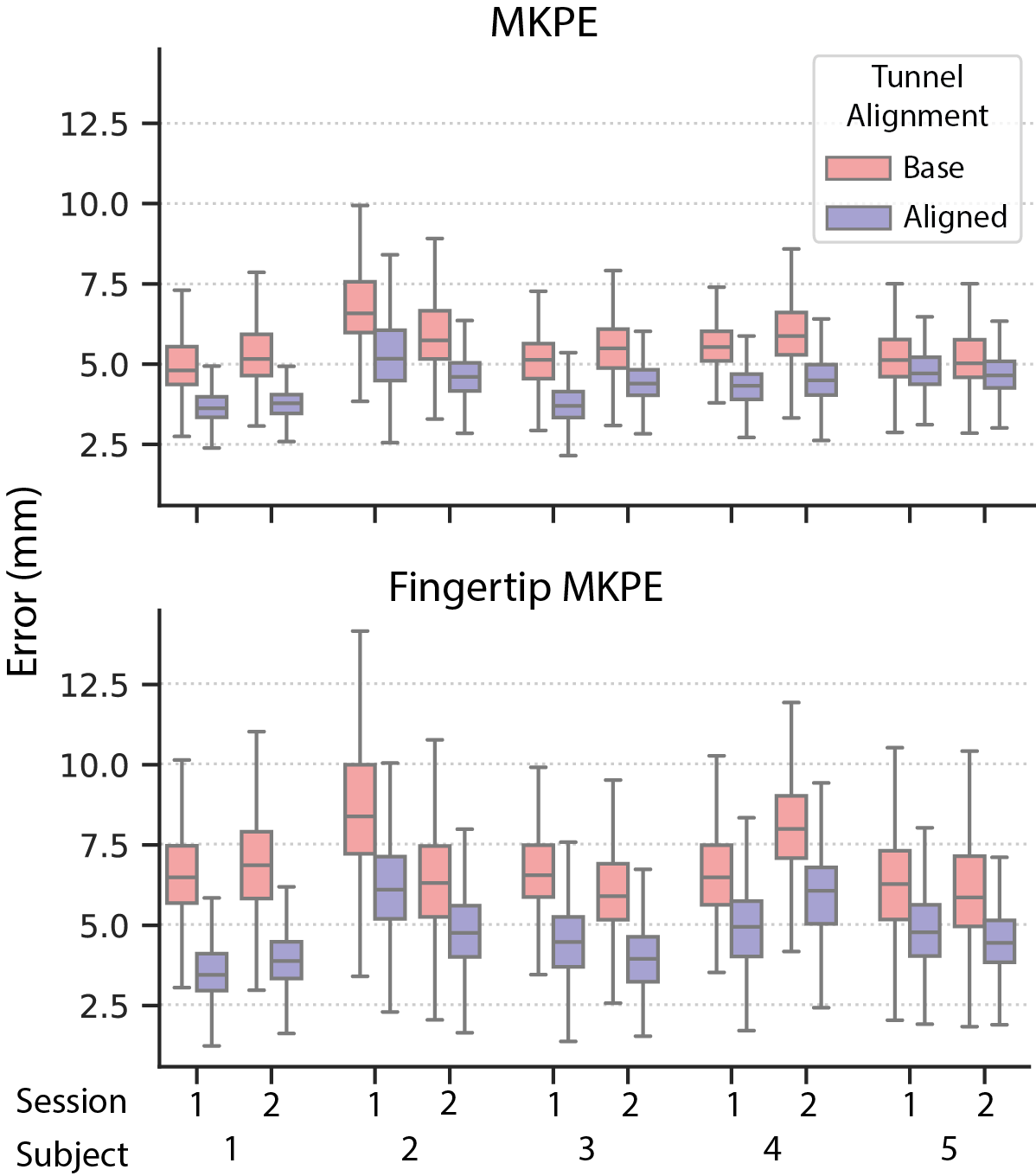}
    \caption{Histogram of landmark errors across subject and sessions. Routing tunnel alignment decreases fingertip error in all cases. After alignment, errors at the fingertips are generally below 5 mm.}
    \label{fig:histogram}
\end{figure}

\begin{table}
\centering
\caption{Per-Fingertip Pose Errors}
\label{tab:per_finger}
\begin{tabular}{lccccc}
\toprule
F.MKPE (mm) & Thumb & Index & Middle & Ring & Pinky \\
\midrule
Base & 9.3 & 5.0 & 7.1 & 5.8 & 9.4 \\
Tunnel Aligned & 7.2 & 4.0 & 4.0 & 3.8 & 6.2 \\
\bottomrule
\end{tabular}
\vspace{-5mm}
\end{table}

\section{DISCUSSION}
We present, to our knowledge, the first method for full hand pose estimation using fiber optic shape sensing with multi-core FBGs. Across a range of object manipulation tasks, our system tracks hand motion with high accuracy (F.MKPE $<$5 mm), positioning the glove as a practical tool for capturing complex dexterous behavior that vision-based trackers often fail to recover due to occlusion (Fig. \ref{fig:occlusion}). Notably, we found that 34\% of mocap frames had a partial occlusion during the object manipulation phase. This is particularly relevant for collecting demonstration data for robot learning, where the most informative signals may be subtle, contact-rich finger adjustments that are frequently hidden from external cameras. We also demonstrate a second application: real-time bimanual teleoperation in virtual reality using tracked hand roots combined with our glove's per-finger pose estimates (Fig. \ref{fig:splash}). This approach could in the future be retargeted to robotic teleoperation use cases.

The low tracking errors are consistent with the high sensitivity of multi-core FBG sensing and its inherent robustness to occlusion and external disturbances. In our experiments, the dominant residual error source was fiber-to-hub shape registration; we found that applying a simple one-time mocap-aided tunnel alignment to correct systematic registration bias reduced fingertip error by roughly 2 mm (Fig. \ref{fig:calibration}, Tables \ref{tab:benchmark}, \ref{tab:per_finger}). Notably, the estimated alignment was stable across sessions and participants, suggesting that a one-time mocap-aided factory calibration may be sufficient to achieve high-accuracy tracking in subsequent captures.

It is possible to use higher end FBG sensing techniques to further reduce fingertip error. For instance, Optical Frequency Domain Reflectometry (OFDR) interrogators facilitate dense FBG sensor spacing as low as 0.3 mm. For this work, we determined that 10 mm sensor spacing was sufficient for our fingertip accuracy requirements ($<$ 5 mm). With either WDM or OFDR approaches, error accumulates along the length of the fiber from the spatial integration of local curvature into 3D shape.

Other sources of error likely arise from both modeling assumptions and implementation details. First, the \emph{virtual hand model} does not perfectly match each user’s true anatomy. Small discrepancies in finger segment lengths or mesh fidelity directly affect fingertip position estimates. Personalized hand models would likely reduce this component of error. Notably, qualitative inspection suggests that each user’s errors were relatively consistent across don and doff conditions (Fig. \ref{fig:histogram}), which supports the interpretation that model mismatch is a systematic source of error.
Second, the inverse kinematics formulation assumes that fibers are routed medially along each finger. In practice, the sensors can shift within the fabric, and the fabric itself can move relative to the underlying skin and bones. These effects are not captured by the current model and introduce additional discrepancies between measured and assumed fiber geometry.
Finally, \emph{hand IK solvers and shape registration} contribute residual error. Although the optimization procedures converged reliably, small residuals remained, indicating that the solutions are approximate rather than exact.

To our knowledge, this is the first FBG glove to reconstruct full hand pose with a hand model. In contrast, previous solutions have used only single core FBGs to directly measure joint flexion at a limited number of finger joints \cite{hu_fiber_2023,rao_study_2023,sun_wearable_2024}. Thus, we cannot directly compare our methods to other FBG gloves, as they measure different errors. We do not compare joint angular error for two reasons. First, many human-to-robot retargeting methods use fingertip positions \cite{meattini_human_2023}. Second, unlike joint angles, whose errors compound through the kinematic chain, fingertip error directly captures the task-relevant deviation at the end-effector.

Other sensorized gloves have evaluated fingertip tracking accuracy. Park et al.\cite{park_stretchable_2024} introduced a stretch-sensing glove with comparable fingertip error (4.02 mm). However, their evaluation uses a much simpler protocol---tracing a 2 cm square---rather than the dexterous manipulation tasks in our study, so direct comparison is not possible.

FSGlove\cite{li_fsglove_2025}, an IMU-based approach, reports fingertip errors of 15.7 mm during a pinching task. This study also evaluated video tracking (Meta Quest 3) and commercial gloves such as MANUS and VRTRIX \cite{noauthor_manus_nodate,noauthor_vrtrix_nodate}, finding lower accuracy than their approach. Although their task and metric are not directly comparable to ours, these results provide a useful reference for error magnitudes from prior custom and commercial gloves. In our experiments (Table \ref{tab:per_finger}), we observe per-finger fingertip errors in a similar range under dexterous manipulation, suggesting that our method is competitive in accuracy for its intended use case. We defer any definitive head-to-head claims to future work under matched conditions. We note that compared to previous FBG glove studies, our inclusion of a hand model allows for future inference into contact geometry, another important area of study in manipulation.

Our method strikes a balance between accuracy, always-available tracking, and portability. We note several improvements and limitations below.
\emph{Mocap-aided calibration} yields our best-accuracy results through a one-time rigid tunnel alignment that corrects systematic registration bias (Section~\ref{subsec:optitrack_cal}), introducing a dependence on external motion capture. This dependence is, however, a one-time rather than a per-capture requirement. In a preliminary analysis across 4 participants, a tunnel alignment estimated from a \emph{different} participant and session reduced fingertip error nearly as much as a subject- and session-matched alignment, with the two differing by only 0.1 mm on average. Because the correction transfers across users and sessions with negligible accuracy loss, the alignment can be treated as a one-time factory calibration rather than a per-session or per-user step, substantially weakening the practical dependence on motion capture. Estimating this alignment instead from anatomical landmarks or a known calibration pose, which would remove the mocap dependence entirely, remains an avenue for future work.
\emph{Hand size} in our tests was limited to large and extra-large hands due to available glove size.
Extension to smaller hands introduces tighter bend radii and closer FBG wavelength spacing, which are relevant engineering considerations but not fundamental limitations---spectral peak merging at tighter radii can be addressed through adjusted fiber specifications or more sophisticated reconstruction algorithms.
\emph{Portability} is primarily determined by the size of the interrogator to which the glove is tethered. The ShapeScan 905 system used in this study supports room or building mobility using a rolling cart---suitable for research needs---whereas unrestricted, in-the-wild captures require miniature optical interrogators. Using \emph{hub tunnels} as a design choice makes the glove look bulky and adds a burden on the algorithm to continuously register fibers to a common reference frame. A good balance between tunnel length, curvature, and exit location is needed to satisfy wearability and accuracy requirements. Lastly, the focus of our evaluation is on wrist-locked skeletal pose reconstruction instead of \emph{world-frame evaluation}, which depends on the specific world tracking accessory attached to the glove.

\addtolength{\textheight}{-12cm}   




\section*{ACKNOWLEDGMENT}
We thank Mia Huang, Francesco Marsili, Mike Hurst, Gaige DeHaven, Keagan McCurdy, Simon Baines, and Stuart Dealey for supporting this work.


\bibliography{bibliography}

\end{document}